\documentclass[journal]{IEEEtran}
\usepackage{hyperref}
\usepackage{amsmath,amssymb}
\usepackage{mathrsfs}

\DeclareMathOperator{\E}{\mathbb{E}}

\usepackage[labelsep=period]{caption}
\usepackage[font=small]{caption}
\usepackage{subcaption}
\usepackage{float}

\ifCLASSINFOpdf
\else
\fi
\usepackage{graphicx}
\begin{document}
\graphicspath{ {./images/} }

%
\title{Locally Interpretable One-Class Anomaly Detection\\ for Credit Card Fraud Detection}
%
%
%

\author{Tung-Yu Wu, You-Ting Wang \\ 
        National Taiwain University \\ 
        \texttt{\{b08901133, b08901072\}@ntu.edu.tw}}
\maketitle

\begin{abstract}
For the highly imbalanced credit card fraud detection problem, most existing methods either use data augmentation methods or conventional machine learning models, while anomaly-detection-based approaches are not sufficient. Furthermore, few studies have employed AI interpretability tools to investigate the feature importance of transaction data, which is crucial for the black-box fraud detection module. Considering these two points together, we propose a novel anomaly detection framework for credit card fraud detection as well as a model-explaining module responsible for prediction explanations. The fraud detection model is composed of two deep neural networks, which are trained in an unsupervised and adversarial manner. Precisely, the generator is an AutoEncoder aiming to reconstruct genuine transaction data, while the discriminator is a fully-connected network for fraud detection. The explanation module has three white-box explainers in charge of interpretations of the AutoEncoder, discriminator, and the whole detection model, respectively. Experimental results show the state-of-the-art performances of our fraud detection model on the benchmark dataset compared with baselines. In addition, prediction analyses by three explainers are presented, offering a clear perspective on how each feature of an instance of interest contributes to the final model output. Our code is available at \href{https://github.com/tony10101105/Locally-Interpretable-One-Class-Anomaly-Detection-for-Credit-Card-Fraud-Detection}{https://github.com/tony10101105/Locally-Interpretable-One-Class-Anomaly-Detection-for-Credit-Card-Fraud-Detection}.
\end{abstract}


%
\IEEEpeerreviewmaketitle

\section{Introduction}
%
%
%
%
With the rapid development of online payment, credit card fraud also continues to climb to new heights, which leads to losses totalling billions of US dollars each year [1]. To address this issue, banks and online payment companies turn to anti-fraud systems to detect illegitimate transactions. Among numerous techniques considered, machine learning (ML) and deep learning (DL) are attracting much attention due to their powerful predictive ability as a fraud detection system [2][3]. Generally, for ML/DL-based methods, models' input features are card transaction data such as the identity of card holders and amount of funds, while outputs are confidence scores that form a probability space for determination of genuine/fraudulent transactions.

Credit card fraud detection, however, is not a typical classification task since fraudulent transactions are extremely rare out of all transactions. Specifically, the dataset is highly skewed, which often causes models to perform poorly when encountering rare cases. There have been some works directly applying traditional ML-based methods [4][5][6][7] where a large dataset is inessential. In terms of DL-based approaches, attempts at high quality data augmentation [8] and the use of key features [9][10] have been made, while few studies [11] regard this problem as anomaly detection which naturally avoids the imbalance problem and has larger room for improvement.

Another crucial topic for DL-based credit card fraud detection systems is the model explainability [12], a research hotspot in the ML community. Considering the black-box property of neural networks, it is desirable to give an explanation of why a transaction is detected as fraudulent for official reports in court. Though various techniques for different levels and perspectives of explanations have been proposed [13][14][15][16][17], investigation into their applications of anomaly detection methods are lacking. 

In this paper, we make the first attempt to leverage a novel anomaly detection framework [18] along with a LIME-based [13] explaining module in the realm of credit card fraud detection. This anomaly detection framework is adopted for its promising performance on detecting irregular images, while the LIME is chosen because of its focus on a single instance of interest which might be the fraudulent transaction in this case. The detection module is adjusted to match the properties of tabular financial data and experiments on comparison with other baseline methods are carried out. LIME explainers are displayed in different parts of the detection module to monitor and inspect the whole prediction pipeline. In summary, the contributions of this work is twofold:
\begin{enumerate}
    \item Modifying and utilizing a novel yet simple anomaly detection architecture to cope with the credit card fraud detection problem, achieving state-of-the-art performance in comparison with other cutting-edge or iconic methods.
    \item Applying LIME to give explanations to different input-output relations for a certain transaction of interest in the detection module.
\end{enumerate}

\section{Related Work}
This paper puts effort into the imbalance issue of credit card fraud detection dataset and the interpretability of utilized anomaly detection framework. In this section, we shall discuss several benchmark methods for imbalanced classification and explainable AI.
\subsection{Imbalanced Classification}
Datasets for detection of fraudulent transactions are inherently skewed because normal cases significantly outnumber dishonest ones. The class imbalance problem greatly increases the difficulty of neural network training. There are two main solutions: (1) manipulating data to balance the dataset and (2) using algorithms like weighted loss function or one-class classification (OCC).

The first approach includes undersampling of majority class and oversampling of minority class. The former deletes excessive data points and the latter generates synthetic data that have similar distributions of real data in minority class, which is much more complicated yet widely adopted. For data synthesis, SMOTE [19] is the most common technique which utilizes k-nearest neighbors algorithm to produce samples close to the minor-class data in the feature space. GAN-based [20] upsampling models [8] have also been broadly studied due to its impressive performance in generating vivid images.

In terms of algorithm-based approaches, weighting of majority and minority class is usually implemented by weighted loss functions. Among them, Focal Loss [21] is one of the most iconic works which encourages networks to down-weight easy examples and focus on learning hard negative examples. One-class classification, also known as anomaly detection from application point of view, merely uses the majority class during training. Conventional ML anomaly detection methods mainly include linear models, such as OCSVM [22], proximity-based models, such as OCNN [23], and probabilistic models, such as COPOD [24]. OCSVM is SVM trained in an unsupervised manner, trying to find a hyperplane that well-encloses the only class in the training set and excludes potential outliers. OCNN is the one-class k-nearest neighbors method which calculates the averaged distance to k nearest data points as the outlier score. COPOD is a copula-based non-parametric statistical approach that can directly provide some decision explanations.

AnoGAN [25] first utilizes adversarial training, where the generator learns to produce samples with distributions similar to major-class data (inliers). As a result, the discriminator could tell whether the input is from a minor class (outliers) by comparing its distribution with that of generated samples. OCAN [11] uses a novel LSTM-AutoEncoder for input feature representation and then conducts the prediction process similar to AnoGAN. These two models both apply a generator to produce fake features sampling from a noise. However, as an intrinsic defect of GAN, the upsampling generation process is unstable. Reference [18] replaces the original noise-to-sample generator with an AutoEncoder for sample reconstruction, which is adopted by many following works [26][27]. For example, Both the GANomaly [26] and Skip-GANomaly [27] have an AutoEncoder as parts of the generator. The former introduces an extra encoder for latent space restriction, while the latter features the use of skip-connections.

\subsection{Explainable AI}
Explainable AI (XAI) is an emerging field in the AI community which aims at allowing human users to comprehend and trust model outputs. Though this domain is relatively new, broad and comprehensive studies [12] have been done. This paper categorizes these methods based on the scope of explanations: (1) local interpretability and (2) global interpretability. 

Local interpretability refers to explanations of decisions a model makes about an instance. Methods in this category either visualize or plot the importance of each input feature to the final output. The saliency map [14], which has long been present and used in local explanations, is a gradient-based visualization approach that calculates each pixel's attribution. LIME [13] proposes a concrete framework for implementations of local surrogate models. Given an instance of interest and a dataset, LIME first generates new data points by perturbing the dataset and getting the black-box predictions for these new data points. Subsequently, a simple transparent model, such as a decision tree, is trained on this dataset, which then should be a good approximation of the target model around this instance. Therefore, analysis on this explainable proxy model could be easily carried out to provide human users with a clear view on how each input feature of this particular instance leads to final outputs made by the black-box model.

Global interpretability answers how model parameters affect the final prediction. It explains the whole model rather than a certain input instance. Activation Maximization (AM) methods [15][16] put forward visualization of neural networks, which is done by finding the input probing image that maximally activates certain neurons with gradient ascent. In particular, reference [16] proposes to use generative models, where the generated images are fed into the target model for classification. We could then examine what kinds of noises and corresponding generated images lead to a certain prediction result. Following works [17] have achieved image-level fine-grained visual explanations for convolutional neural networks (CNN).

\begin{figure}[t]
    \centering
    \includegraphics[width=0.45\textwidth]{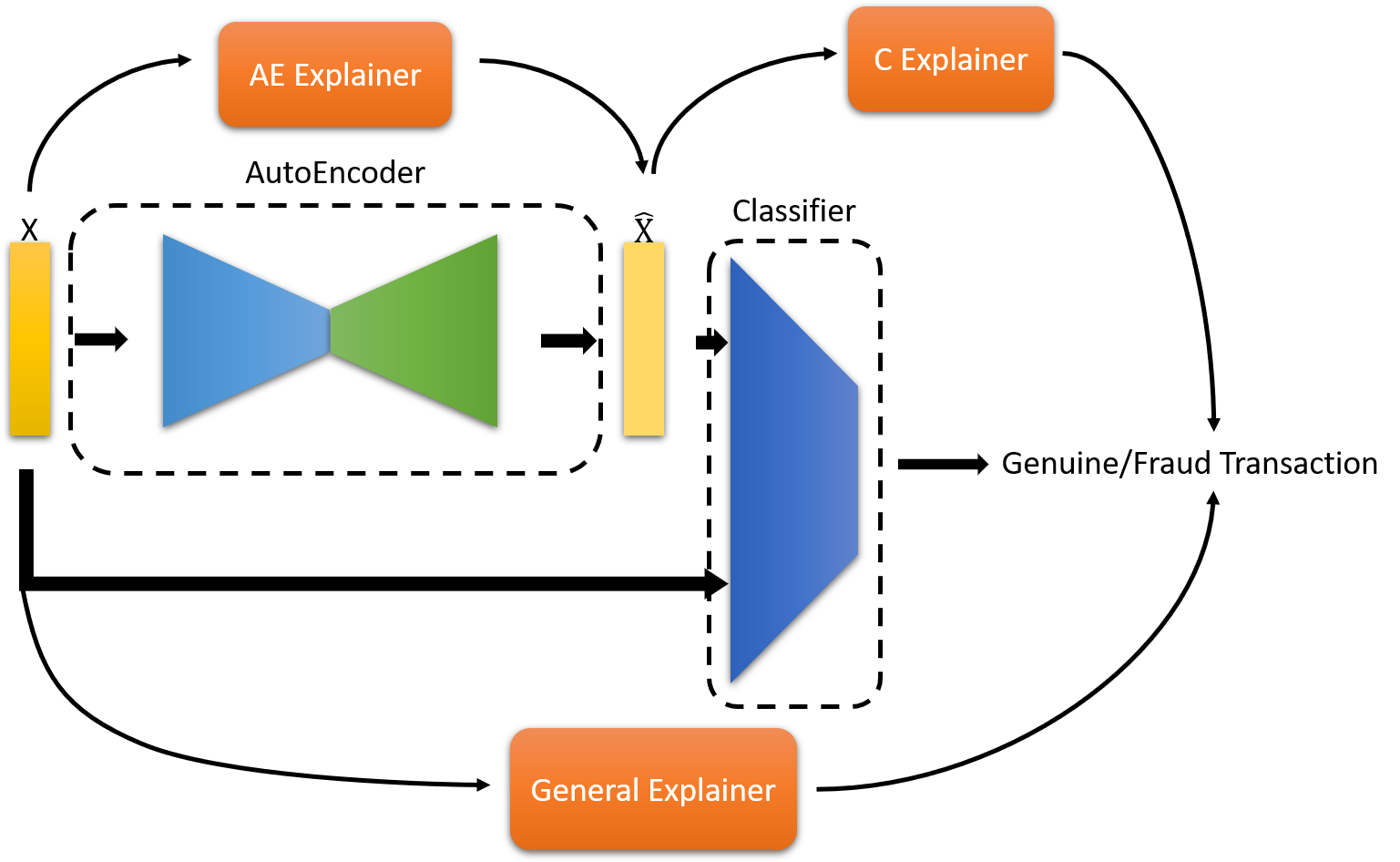}
    \caption{Structure of our method.}
\end{figure}

\section{Methodology}
The proposed framework comprises two modules: (1) the anomaly detection model and (2) the model explainers. The former's architecture is mainly derived from [18], which consists of an AutoEncoder for input-output reconstruction and a fully-connected classifier for fraud detection. The two networks are trained in an adversarial and unsupervised manner so as to cope with the class imbalance issue, and some adjustments are applied on account of specific properties of this credit card fraud detection task. The model explainers provide users with a clear perspective on how the network output is influenced by input features. We choose LIME to analyze a single transaction instance of interest. The overview of the proposed structure is shown in Fig 1.

\subsection{Anomaly Detection Model - AutoEncoder}
Capable of reconstructing and denoising input samples [18][26][27], AutoEncoders have been widely used in novelty detection tasks recently. In this paper, the target and non-target class are the genuine and fraudulent transaction, respectively. For the low-dimensional tabular financial transaction data, we adopt an AutoEncoder for reconstruction. That is, no extra Gaussian noise is added in the input.

The model is composed of fully-connected layers. Batch Normalization (BN) and Rectified Linear Unit (ReLU) are utilized to ensure stable gradient flows. Given that fraudulent transactions are extremely rare, we only train the AutoEncoder with genuine transactions. Specifically, the model is trained to preserve the distribution of target-class inputs, while the outliers, namely, non-target class, will be naturally mapped to an uncertain point in the latent space and fail to be reconstructed due to their absence from the training set. To achieve this goal, two loss functions are presented:
\begin{equation}
    L_\mathcal{R} = \|\mathcal{R}(X) - X)\|_2,
\end{equation}
\begin{equation}
    L_\mathcal{R}^{GAN} = -log{\ \mathcal{C}(\mathcal{R}(X))}.
\end{equation}
Equation (1) is the reconstruction loss, which is decided to be L2 loss in the next section. R denotes the reconstructor, that is, the AutoEncoder. X is the transaction data. Equation (2), which is a part of adversarial loss, is the binary cross entropy that guides the AutoEncoder to produce outputs that have the same distribution as inputs and subsequently confuse the discriminator. C stands for the Classifier. The overall training objective of the AutoEncoder is then: 
\begin{equation}
    \min(\E_{X \sim pt}[\|\mathcal{R}(X) - X)\|_2 -log{\ \mathcal{C}(\mathcal{R}(X))}]),
\end{equation}
where pt denotes the distribution of genuine transactions.

\subsection{Anomaly Detection Model - Classifier}
The classifier is a simple Multilayer Perceptron (MLP) that aims to separate original transaction data from the reconstructed one. The output of this model is a single value representing the probability of target class. As a result, the binary cross entropy loss for the Classifier can be as follows: 
\begin{equation}
    L_{\mathcal{C}}^{GAN} = -log{\ \mathcal{C}(X)} - log{(1- \mathcal{C}(\mathcal{R}(X)))}.
\end{equation}
The adversarial training objective in the whole anomaly detection model can then be summarized as:
\begin{equation}
    \begin{split}
        \min_{\mathcal{R}}\max_{\mathcal{C}}(\E_{X \sim pt}[log{\ \mathcal{C}(X)} + log{(1- \mathcal{C}(\mathcal{R}(X)))}]).
    \end{split}
\end{equation}
It is notable that the training procedure is done when the AutoEncoder and classifier reach the Nash equilibrium and the reconstruction loss is small. This means the reconstruction process is performed well and the classifier outputs around 0.5 on both original and reconstructed features, which indicates that their distributions are too close to be distinguished. In this situation, it is simple and straightforward to determine by a threshold whether the instance of interest is a fraudulent transaction, which is mapped to an unknown distribution when passing through the AutoEncoder and leads to a prediction value far from 0.5 at the classifier. The threshold is set to be 0.7 in this work.

\subsection{Model Explainer}
In terms of the explainability of our one-class fraud detection framework, we are particularly interested in the interpretations of (1) how the reconstruction stage is affected by genuine/fraudulent transaction data, (2) how the consequent reconstructed features bring about final prediction results, and (3) the overall credit card fraud detection module. Therefore, three LIME-based explainers, which are simple white-box models, are employed to interpret the three different input-output relations. According to their functions, models are named as AE explainer, C explainer, and general explainer, respectively, as demonstrated in Fig 1.

As a model-agnostic technique, LIME only requires the target model to be a classifier or regressor. Thus, the C and general classification explainer can be directly trained with samples acquired by sampling from the distribution of the given dataset, while the vector output of AutoEncoder obstructs this way for the AE explainer. To address this problem, we make use of L2-norm to compute a single value of reconstruction error as the label for the training of the AE regression explainer. This is equivalent to (1), while we describe the formula here to clarify the calculation of label:
\begin{equation}
    label = \frac{\sum\limits_{i = 1}^n{(\mathcal{R}(X)_i - X_i)^2}}{n},
\end{equation}
where the character n is the number of features. With this step, the AE explainer is turned to an explainable regression model whose outputs denote L2 reconstruction error. 

\setlength{\arrayrulewidth}{0.5pt}
\setlength{\tabcolsep}{25pt}
\renewcommand{\arraystretch}{1.5}

\begin{table*}[t]
    \centering
    \caption{Credit card fraud detection results on accuracy, precision, recall, F1-score, and Matthews correlation coefficient}
    \begin{tabular}{c c c c c c}
    \hline
    Methods & Accuracy & Precision & Recall & F1-score & MCC \\
    \hline
    OCSVM & 0.8898 & 0.9204 & 0.8673 & 0.8931 & 0.7811 \\
    OCNN & 0.9000 & 0.9122 & 0.8904 & 0.9012 & 0.8002 \\
    COPOD & 0.8388 & 0.7224 & 0.9415 & 0.8175 & 0.6967 \\
    AutoEncoder & 0.8000 & \textbf{0.9245} & 0.7402 & 0.8221 & 0.6195 \\
    OCAN & 0.8806 & 0.8061 & \textbf{0.9472} & 0.8710 & 0.7698 \\
    Ours & \textbf{0.9061} & 0.9216 & 0.8878 & \textbf{0.9044} & \textbf{0.8128} \\
    \hline
    \end{tabular}
\end{table*}

\section{Experiments}
\subsection{Experimental setup}
\subsubsection{Datasets}
We test the performance of our anomaly detection framework on a benchmark credit card fraud detection dataset\footnote{https://www.kaggle.com/mlg-ulb/creditcardfraud}. The dataset contains 284,807 credit card transactions, collected in Europe during a 2-day period in September 2013. There are only 492 fraudulent cases (minority class, labeled as 1) accounting for 0.172\% and the other 234315 cases are genuine transactions (majority class, labeled as 0), which indicates the dataset is highly imbalanced. Each transaction has 30 features, 28 of which are principal components obtained from PCA and the other two are time and amount. The 28 features are listed as V1 to V28 and further information is not provided because of confidentiality issues. Time feature denotes the elapsed time between the current and first transaction, which is considered in temporal-based models [9][10]. Amount feature is the amount of money in the transaction, and it is used in cost-sensitive learning. We only utilize the 28 features in this work as in [11]. Since the data have been normalized before PCA, we do not clean them again before the training process. Since the original dataset is not split into training and testing sets, we select 490 out of 492 fraudulent cases and 490 out of 234315 genuine cases to generate a well-balanced testing set, and the remaining 233825 genuine cases form the training set for our model. All the baselines are tested on this testing set, while training sets for different methods may differ. 

\subsubsection{anomaly detection methods}
We compare our framework with five baseline models: OCSVM, OCNN, COPOD, vanilla AutoEncoder, and OCAN. For the first four baselines, we use implementations provided in PyOD [28], which is a renowned high-quality toolkit for novelty detection. OCAN is not included in this package, so we retrieve its open-source original implementation\footnote{https://github.com/PanpanZheng/OCAN}. In terms of training parameters, OCSVM, COPOD, AutoEncoder, and OCAN are trained with 700 genuine cases. OCNN has no training stage, but an evaluation set is required for hyper-parameter tuning. Thus, we randomly pick 25 genuine and 25 fraudulent cases for evaluation and subsequently determine k = 5. All hyper-parameters are manually adjusted to reach the best performances if no default values are provided by the original paper or source code.

Our model is trained on the whole training set for 2 epochs. The batch size is 4096, and the learning rate is 2e-4. Weight initialization is applied to ensure a fine starting point. The threshold is 0.7, and a classifier output above this value is identified as an illegal transaction.

\subsubsection{LIME}
We use LIME's open-source implementations\footnote{https://github.com/marcotcr/lime} to build the three explainers. The AE explainer is a regression model, where the label is defined in (6). The D and general explainer are classifiers, where the label is the output of the fraud detection module.

Firstly, we select a single fraudulent case from the testing set. Secondly, 5000 data points are produced by sampling from the distribution of the class-balanced testing set, and their corresponding outputs by our anomaly detection model are used as labels. Finally, these data points and labels serve as the training set to build a simple explainable model, into which we can feed the original fraudulent case to obtain its prediction explanation.

\subsection{Results}
\subsubsection{Anomaly Detection Performance}
The performance measures include accuracy, precision, recall, F1-score, and Matthews correlation coefficient (MCC). Particularly, F1-score and MCC are balanced binary classification measures that take the entire confusion matrix into account. Precision indicates how likely a case detected asf fraudulent is indeed illegitimate. Recall denotes the proportion of fraudulent cases that are successfully detected. The testing results are displayed in Table 1.

It is remarkable that our method achieves 0.9061 accuracy, 0.9044 F1-score, and 0.8128 MCC, higher than any other baselines. This indicates that our model reaches the best performance if we consider precision and recall simultaneously. For precision only, AutoEncoder sacrifices recall to reach the highest precision (0.9245), resulting in a lower accuracy, F1-score, and MCC. However, this merely outperforms our result (0.9216), by a small margin. For recall only, OCAN and COPOD obtain promising results (0.9472 and 0.9415, respectively). Nevertheless, they fail to reach a balance between precision and recall. If we compare OCSVM, OCNN, and our method, which have relatively balanced precision and recall, our method almost outperforms the other two models in all indices except for recall, where OCNN is only slightly better.

We also compare different loss functions to determine the most suitable form for reconstruction. Three candidates are L1, SmoothL1, and L2 loss. For simplicity, we finally choose L2 loss because it results in the best MCC, as shown in Table 2.

\begin{table}[h]
    \centering
    \caption{MCC of different reconstruction losses}
    \begin{tabular}{ | c | c | }
        \hline
           Loss Function & MCC \\ 
        \hline
        L1  & 0.7990 \\ 
        \hline
        SmoothL1 & 0.7970 \\
        \hline
        L2  & \textbf{0.8128} \\ 
        \hline
    \end{tabular}
\end{table}

Finally, we plot the receiver operating characteristic (ROC) curve and calculate the area under curve (AUC) in Fig 2. ROC and AUC show the classifier performance at various thresholds and serve as a criterion for model robustness. The AUC value ranges from 0 to 1. Specifically, 0.5 (such as the diagonal in Fig 2) denotes a naive classifier and 1 indicates a perfect classifier. Our method achieves an AUC of 0.9434.

\begin{figure}[H]
    \centering
    \includegraphics[width=0.3\textwidth]{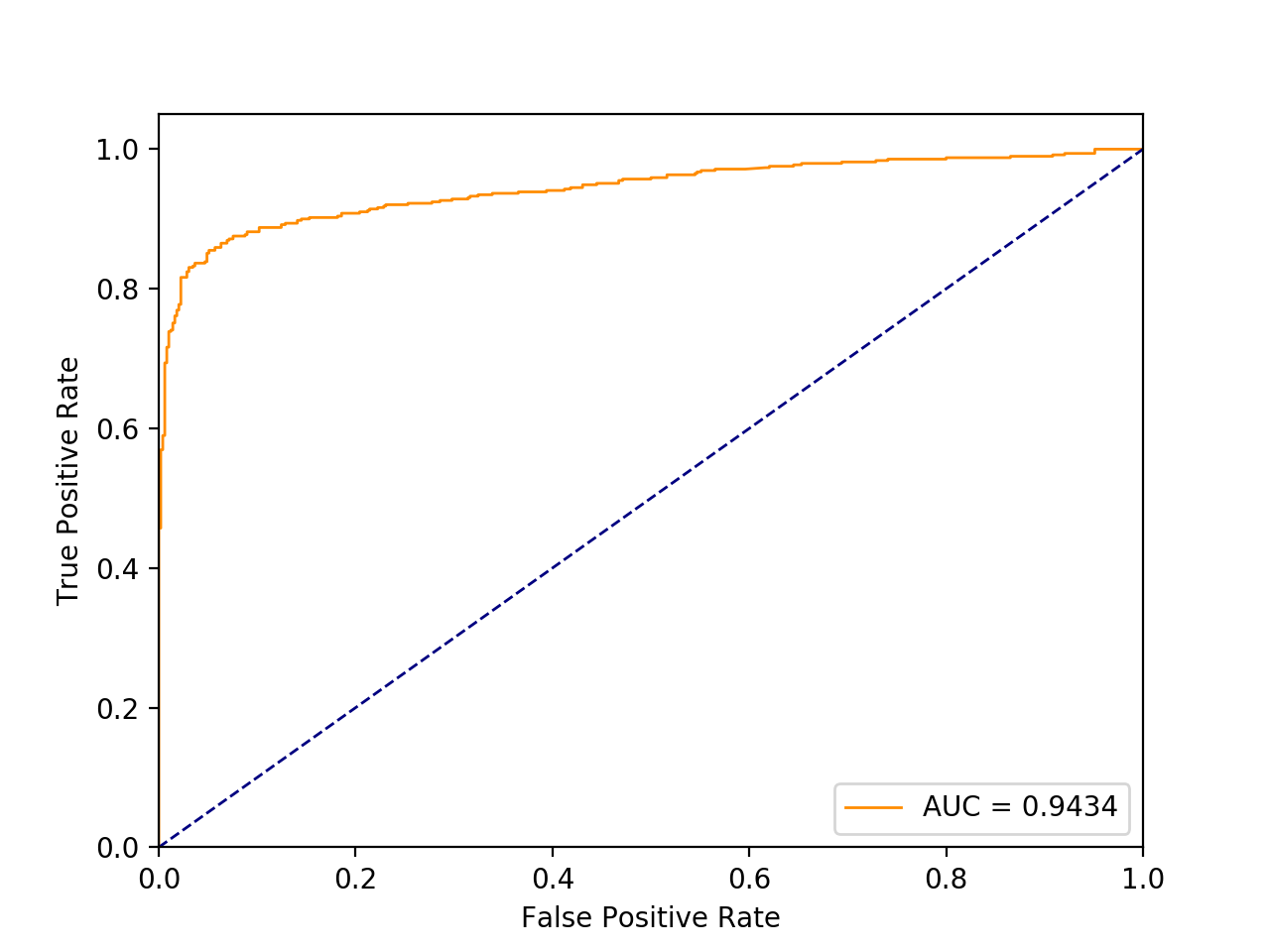}
    \caption{ROC and AUC of our model}
\end{figure}

\subsubsection{Model Interpretation}
Fig 3., 4., and 5. display interpretations of a fraudulent transaction by the AE, C, and general explainers, respectively. The left sub-figure of each figure (Fig 3(a), 4(a), 5(a)) is the output value of the explainer. The middle bar chart (Fig 3(b), 4(b), 5(b)) shows the contribution of each feature value to the explainer output. The right table (Fig 3(c), 4(c), 5(c)) lists down the value of each feature. For clarity, we only plot the top six important features out of 28 features.

\begin{figure}[H]
     \centering
     \begin{subfigure}[b]{0.17\textwidth}
         \centering
         \includegraphics[width=\textwidth]{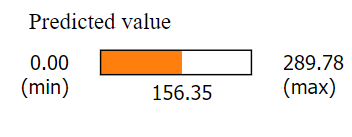}
         \caption{}
     \end{subfigure}
     \hfill
     \begin{subfigure}[b]{0.2\textwidth}
         \centering
         \includegraphics[width=\textwidth]{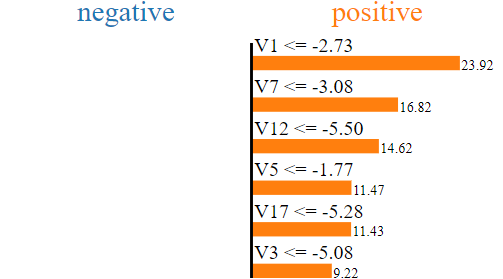}
         \caption{}
     \end{subfigure}
     \hfill
     \begin{subfigure}[b]{0.08\textwidth}
         \centering
         \includegraphics[width=\textwidth]{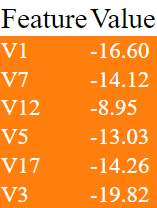}
         \caption{}
     \end{subfigure}
        \caption{Interpretations by the AE explainer}
        \label{fig:three graphs}
\end{figure}

We first elaborate on Fig 3. Since the input transaction is fraudulent, it is mapped to an unknown distribution rather than being perfectly reconstructed, as discussed in the previous section. This leads to a large reconstructed error up to 156.35. Then, by disturbing the fraudulent input, it is found that V1 possesses the highest influence among all features. If V1 rises from -16.60, as listed in the right feature value table, to -2.73, as shown in the middle chart, the predicted value will be reduced to 156.35 - 23.92 = 132.43. 

\begin{figure}[H]
     \centering
     \begin{subfigure}[b]{0.17\textwidth}
         \centering
         \includegraphics[width=\textwidth]{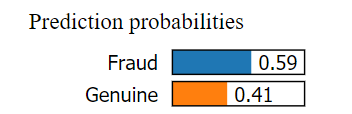}
         \caption{}
     \end{subfigure}
     \hfill
     \begin{subfigure}[b]{0.2\textwidth}
         \centering
         \includegraphics[width=\textwidth]{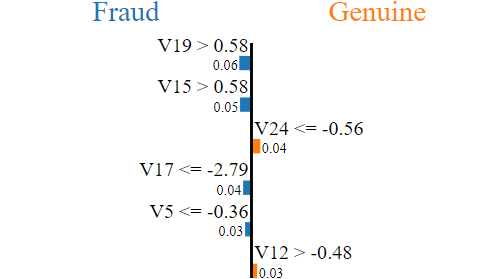}
         \caption{}
     \end{subfigure}
     \hfill
     \begin{subfigure}[b]{0.08\textwidth}
         \centering
         \includegraphics[width=\textwidth]{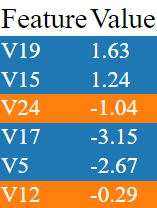}
         \caption{}
     \end{subfigure}
        \caption{Interpretations by the C explainer}
        \label{fig:three graphs}
\end{figure}

Fig 4. shows how each feature of the reconstructed input affects the classifier prediction. If V19 decreases from 1.63 to 0.58, the classifier will reduce its confidence of fraudulent transaction from 0.59 to 0.53. Observing Fig 3(c) and 4(c), we can find that feature values of the reconstructed inputs are significantly different from that of the original input. This matches the high reconstruction error shown in Fig 3.

\begin{figure}[H]
     \centering
     \begin{subfigure}[b]{0.17\textwidth}
         \centering
         \includegraphics[width=\textwidth]{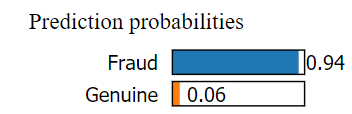}
         \caption{}
     \end{subfigure}
     \hfill
     \begin{subfigure}[b]{0.2\textwidth}
         \centering
         \includegraphics[width=\textwidth]{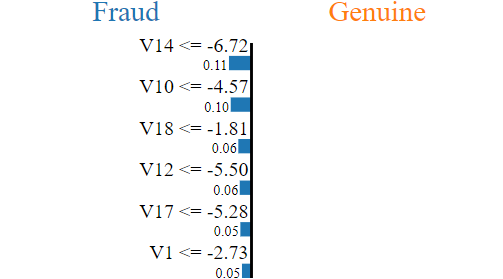}
         \caption{}
     \end{subfigure}
     \hfill
     \begin{subfigure}[b]{0.08\textwidth}
         \centering
         \includegraphics[width=\textwidth]{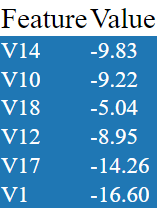}
         \caption{}
     \end{subfigure}
        \caption{Interpretations by the general explainer}
        \label{fig:three graphs}
\end{figure}

Fig 5. is the overall analysis. The explainer is a single transparent model, where inputs are original transactions and outputs are predictions. It approximates the entire fraud detection module and the provided analysis can be interpreted as in Fig 3. and Fig 4.

\section{Conclusion}
This work proposes to leverage an adversarially trained anomaly detection model for credit card fraud detection problem. Experimental results show that this framework outperforms other iconic or state-of-the-art baseline models. Furthermore, LIME is applied to investigate input-output relations of this fraud detection model and analyses of an instance of interest are presented, providing a clear view on how each input feature influences the final prediction. Future work will focus on interpretability of other unsupervised or semi-supervised methods in the application area.


%



\ifCLASSOPTIONcaptionsoff
  \newpage
\fi

\end{document}